\def\year{2019}\relax
\documentclass[letterpaper]{article} %DO NOT CHANGE THIS
\usepackage{aaai19}  %Required
\usepackage{times}  %Required
\usepackage{helvet}  %Required
\usepackage{courier}  %Required
\usepackage{url}  %Required
\usepackage{graphicx}  %Required
\usepackage{amsmath} % add by Ryo
\usepackage{amsfonts} % add by Ryo
\usepackage{fontawesome} % add by Ryo
\begin{document}
\title{Another Diversity-Promoting Objective Function for Neural Dialogue Generation}
\author{Ryo Nakamura, Katsuhito Sudoh, Koichiro Yoshino, Satoshi Nakamura\\
Graduate School of Information Science, Nara Institute of Science and Techonology, Japan\\
RIKEN, Center for Advanced Intelligence Project AIP, Japan\\
\{nakamura.ryo.nm8, sudoh, koichiro, s-nakamura\}@is.naist.jp\\
}
\maketitle
\begin{abstract}
Although generation-based dialogue systems have been widely researched, the response generations by most existing systems have very low diversities.
The most likely reason for this problem is Maximum Likelihood Estimation (MLE) with Softmax Cross-Entropy (SCE) loss.
MLE trains models to generate the most frequent responses from enormous generation candidates, although in actual dialogues there are various responses based on the context.
In this paper, we propose a new objective function called Inverse Token Frequency (ITF) loss, which individually scales smaller loss for frequent token classes and larger loss for rare token classes.
This function encourages the model to generate rare tokens rather than frequent tokens.
It does not complicate the model and its training is stable because we only replace the objective function.
On the OpenSubtitles dialogue dataset, our loss model establishes a state-of-the-art DIST-1 of 7.56, which is the unigram diversity score, while maintaining a good BLEU-1 score.
On a Japanese Twitter replies dataset, our loss model achieves a DIST-1 score comparable to the ground truth.
\end{abstract}

\section{Introduction}

Researchers have widely investigated generation-based dialogue systems and are making rapid progress in this area.
However, a common problem remains: dialogue systems tend to produce such generic responses as ``I don't know.''
Some studies have artificially promoted diversity.
A diversity-promoting objective function based on Maximum Mutual Information (MMI) first addressed this kind of problem \cite{li2016diversity}, and various generative model-based methods (e.g., GAN and VAE) have been proposed \cite{li2017adversarial,xu2018dp,olabiyi2018multi,cao2017latent,zhao2017learning}.

%%%

The most likely reason for this problem is Maximum Likelihood Estimation (MLE) with Softmax Cross-Entropy (SCE) loss.
Although many different responses exist in actual dialogues, when you are talking to a human, MLE trains the model to generate frequent phrases in the training set, such as ``I'm sorry,'' ``I'm not sure,'' and ``I don't know.''

%%%

To solve this low diversity problem, we propose a new objective function called Inverse Token Frequency (ITF) loss, which scales loss based on the ITF at each time step.
This new function encourages the model to generate rare tokens rather than frequent tokens.
ITF loss creates the following advantages:

\begin{itemize}
\item The ITF loss model yields state-of-the-art diversity and maintains the quality. It is also very clear, easy to understand, and sufficiently novel.
\item ITF loss can be easily incorporated, and the  MMI, GAN, VAE, and RL implementations become complicated because models are modified or added.
\item Training with ITF loss is as stable as training with MLE, and training with GAN and RL is usually unstable and often requires pre-training with MLE. 
\end{itemize}

\begin{table*}[t]
  \begin{center}
    \caption{Examples of token frequencies and corresponding weights ($w$) with $\lambda= 0.4$ on English OpenSubtitles dialogue. We tokenized sentences by a subword model with a 32k vocabulary using Sentencepiece \cite{kudo2018subword}, and an underscore (\_) stands for a word boundary given by Sentencepiece.}
    \begin{tabular}{cccc||cccc||cccc}
      \\
      index & token & freq & $w$ & index & token & freq & $w$ & index & token & freq & $w$ \\ \hline
      10 & \_I & 1096434 & 0.00384 & 1000 & \_strong & 2872 & 0.0414 & 10000 & \_cupboard & 186 & 0.124 \\
      20 & \_it & 383979 & 0.00584 & 2000 & But & 1281 & 0.0571 & 15000 & \_cruelty & 107 & 0.154 \\
      50 & \_don & 128837 & 0.00904 & 3000 & rd & 795 & 0.0692 & 20000 & TOO & 69 & 0.184 \\
      100 & \_them & 54395 & 0.0128 & 4000 & \_print & 559 & 0.0796 & 25000 & \_planetarium & 46 & 0.216 \\
      500 & \_happen & 7040 & 0.0289 & 5000 & \_bottles & 425 & 0.0888 & 30000 & ebulon & 29 & 0.260 \\
    \end{tabular}
    \label{tab:w}
  \end{center}
\end{table*}

\section{Related Works}

Low diversity problems in neural dialogue generation were first addressed by \citeauthor{li2016diversity} (\citeyear{li2016diversity}) who augmented the objective function with Maximum Mutual Information (MMI).
Their work promoted diversity by penalizing generic responses with an anti-language model.
For sustainable dialogue generation, a reinforcement learning-based method was proposed by \citeauthor{li2016deep} (\citeyear{li2016deep}).
The negative cosine similarity between an input and a response was given as a reward, but the improvement of the diversity was small.
Controlling output tokens by attention or an extension to LSTM cells leads to the diversity of response generation \cite{wen2015semantically,zhou2017mechanism,shao2017generating}.
Encoding dialog histories and external resources also promoted diversity \cite{ghazvininejad2017knowledge}.
Even though other works addressed over-generation and reranking \cite{wen2015stochastic,li2016diversity,serban2017deep}, a model must be built that can generate a variety of sentences.

%%%

Recently, several generative model-based methods have been proposed.
The Generative Adversarial Network (GAN) was proposed in image generation \cite{goodfellow2014generative} and applied to text generation \cite{yu2017seqgan} and dialogue generation \cite{li2017adversarial,xu2018dp,olabiyi2018multi}.
Currently, training with GAN for dialogue generation is very unstable and requires pre-training.
Variational AutoEncoder (VAE) was also proposed in image generation \cite{kingma2013auto} and applied to text generation \cite{bowman2016generating} and dialogue generation \cite{cao2017latent,zhao2017learning}.

\section{Methods}

The task of response generation can be formulated as a sequence-to-sequence problem that generates a response based on given inputs.
In neural dialogue generation, training with Maximum Likelihood Estimation (MLE) approximates model distribution that generates response sentence $Y = \{y_1, y_2, \ldots, y_m\}$ to a true distribution that gives target sentence $T = \{t_1, t_2, \ldots, t_m\}$.
Generally, the loss function individually calculates the loss between generated token $y_i$ and target token $t_i$ across all token symbols.
The following section describes the loss at any time step $i$.

\subsection{Softmax Cross-Entropy Loss}

Softmax Cross-Entropy (SCE) loss, which is commonly used when training a sequence-to-sequence model with MLE, is typically defined as:

\begin{equation}
  L_{{\rm sce}} = - \log\left(\frac{\exp(x_c)}{\sum_k^{|V|} \exp(x_k)}\right),
\end{equation}

\noindent where $x_k$ is the $k$-th element of $x \in \mathbb{R}^{|V|}$, which is the output of the projection layer before the softmax layer, and $c$ is the index of the target token class.
SCE loss treats each token class equally.
Therefore, the generation probabilities of the frequent tokens become too large, and those of the rare tokens become too small.
This problem causes the model to select only frequent tokens from an enormous number of token candidates.

\subsection{Inverse Token Frequency Loss}

We propose Inverse Token Frequency (ITF) loss to deal with the bias of SCE loss and to promote diversity.
ITF loss is a frequency-weighted version of SCE loss:

\begin{eqnarray}
  L_{{\rm ITF}} &=& w_c L_{{\rm sce}} \\
  w_c &=& \frac{1}{{\rm freq}({\rm token}_c)^{\lambda}},
\label{eq:weight}
\end{eqnarray}

\noindent where $w_c$ is an element corresponding to class $c$ in weight $w \in \mathbb{R}^{|V|}$, ${\rm token}_c$ is a token corresponding to class $c$, and function ${\rm freq}({\rm token}_c)$ is the frequency with which ${\rm token}_c$ appears in the training set.
Hyperparameter $\lambda$ controls the frequency's impact.
When $\lambda = 0$, the ITF loss is equivalent to the SCE loss.
The distribution drawn from the softmax layer is the same for training and evaluation.
Special tokens, such as Start and End (i.e., starting and ending of sentences), are handled identically as the others.
Therefore, these tokens are very small weight in the ITF loss because they appear in all of the sentences in the training set.
We found no serious problems, including generating inappropriately long responses by weighting.
In Table \ref{tab:w}, we show some examples of token frequencies.

%%%

Finally, we show a code example of ITF loss implementation with PyTorch:

\begin{quote}
\begin{scriptsize}\begin{verbatim}
def get_weights(_lambda):
    weights = torch.zeros(vocab_size)
    for token, index in token2index.items():
        weight = 1 / (token2freq[token]**_lambda)
        weights[index] = weight
    return weights

weights = get_weights(_lambda=0.4)
itf_loss = nn.NLLLoss(weight=weights)
log_softmax = nn.LogSoftmax(dim=-1)

def train_step(...):
    prob = log_softmax(model_output)
    loss = 0
    for i in range(sequence_size):
        loss += itf_loss(prob[i], target[i])
\end{verbatim}\end{scriptsize}
\end{quote}

\section{Experiments}

We experimentally compared the diversity of the dialogue generation  of our ITF loss model and previous methods using three dialogue datasets in a couple of different domains and languages.

\subsection{Training Details}

We chose $\lambda = 0.4$ for all the ITF loss models based on the discussion below in Section \ref{sec:lambda}.
In the decoder, we applied a repetition suppressor with $\lambda = 1$ in all the models to suppress the repetitive generation of identical phrases for improving the quality.
Details are discussed in Section \ref{sec:sup_rep}.

%%%

In all the models, we set four layers in both the encoder and the decoder, 256 hidden units, an embedding size of 256, a maximum sequence size of 28, a mini-batch size of 32 and trained them with the Adam Optimizer at a learning rate of 3e-4.
We tokenized the sentences by a subword model with a 32k vocabulary using Sentencepiece \cite{kudo2018subword}.

\subsection{Baselines}

We compared our loss model to some competitive models.

\subsubsection{Seq2Seq}

An encoder-decoder (a.k.a., sequence-to-sequence) network has been applied to many generation-based dialogue systems \cite{shang2015neural,vinyals2015neural,sordoni2015neural}.
We used it with the bidirectional multi-layered LSTM encoder and the multi-layered LSTM decoder, both of which have residual connections around each layer. The bidirectional LSTM encoder compresses well the feature representation of the whole source sentence, and the residual connection helps train the deep neural networks.

\subsubsection{Seq2Seq + Attn}

An attention mechanism improved the performance and the diversity by referring to encoded memory \cite{zhou2017mechanism,shao2017generating}.
In the above basic Seq2Seq, as the decoding process continues, the constraints from the source sentence often weaken, and then the decoding depends on the generated tokens like in a language model. 
Since the attention mechanism refers to the feature representation of the source sentence at each time step, it helps avoid language model-like generation and increases diversity.
We use the encoder-decoder, which controls the decoder by Scaled Dot-Product Attention \cite{vaswani2017attention}.

\subsubsection{Seq2Seq + MMI}

Based on MMI-antiLM inference \cite{li2016diversity}, the Maximum Mutual Information objective function is defined:

\begin{equation}
  \hat{T} = {\rm argmax} \left\{\log p(T|S) - \lambda\log p(T)\right\},
\end{equation}

\noindent where $\log p(T|S)$ is the conditional log-likelihood of a generated sentence given a source sentence and $\log p(T)$ is the unconditional log-likelihood of the generated sentence as a language model.
By subtracting a language model term, MMI-antiLM suppresses language model-like generation.
Note that the diversity does not improve when MMI-antiLM is used during the training.
As described by \citeauthor{li2016diversity} (\citeyear{li2016diversity}), we used MLE during the training and MMI-antiLM during the evaluation.
In practice, MMI-antiLM generates token $y$:

\begin{equation}
  y = {\rm argmax} \left\{\log {\rm softmax} (x - \lambda u) \right\},
\end{equation}

\noindent where $x \in \mathbb{R}^{|V|}$ is the output of the projection layer using the encoder-decoder given a source sentence and $u \in \mathbb{R}^{|V|}$ is the output of the projection layer using only the decoder (i.e., initial value of the LSTM's hidden state is set to zeros). Other formulations, such as $\log{\rm softmax}(x) - \log{\rm softmax}(\lambda u)$ and $\log({\rm softmax}(x) - {\rm softmax}(\lambda u))$, did not work well in our preliminary experiment.
Coefficient $\lambda$ is the degree of the anti-language model.
We chose $\lambda = 0.8$ for all the datasets and
only applied MMI-antiLM to the first five time steps of the decoder (i.e., $\gamma = 5$).

\subsubsection{MemN2N}

In dialogue generation, models can acquire contextual consistency by referring to multi-turn utterances as a dialogue history.
The Memory Network (MemN2N) and the Hierarchical Recurrent Encoder-Decoder (HRED) are typical ways to encode multiple utterances \cite{sukhbaatar2015end,miller2016key,serban2016building,serban2017hierarchical}.
We use the former, which encodes the dialogue histories of multi-turns, stores them in memory slots, and extracts contextual information by attention.
We generated a sentence vector from a tokens matrix with a bidirectional multi-layered LSTM instead of summation with positional encoding.
We always applied temporal encoding.

\subsection{Evaluation Details}

We used BLEU to measure the quality of the generated sentences and DIST to measure the diversity.
The following are the details of each metric.

\subsubsection{BLEU}

BLEU-n calculates the percentage of n-gram matching between all of the generated sentences and all of the reference sentences \cite{papineni2002bleu}.
We calculated the corpus-level BLEU-1 and BLEU-2 scores that measure the degree of unigram and up to bigram matching.
We also applied a brevity penalty that incorporated recall and a smoothing method that added $0.1$ counts to precision with $0$ counts.

%%%

Some dialogue generation studies obtained BLEU-4 scores, but in our experiments the BLEU-4 scores were very low, typically less than $2$.
Because there are an enormous number of generation candidates, higher-order n-grams are hardly matched in the reference, and scores slide up and down depending on the initializing model and the sampling differences of the mini-batches.
Therefore, the corresponding BLEU-4 scores become more unstable.

\subsubsection{DIST}

DIST-n calculates the percentage of the distinct n-grams in all the n-grams of the generated responses proposed by \cite{li2016diversity}.
We calculated the DIST-1 and DIST-2 scores that measure the degree of the unigram and bigram diversity.

\subsubsection{}

We tokenized with the TweetTokenizer in the NLTK to calculate the BLEU and DIST scores for the word sequences (not subword sequences).
Note that for the Japanese Twitter replies, we tokenized with SentencePiece to directly calculate the BLEU and DIST scores for the subword sequences because no Japanese tokenizer was suitable for tweet data.
We removed such symbols as Padding, Unknown, Start, and End from all the sentences during the evaluations.
Because a beam search maximizes the likelihood of the whole sentence and causes low diversity, the decoders of all the models generate tokens by a greedy search.

\subsection{Results}

\subsubsection{OpenSubtitles}\label{sec:opensubtitles_data}

\begin{table}[t]
  \begin{center}
    \caption{Results of English OpenSubtitles dialogue. BLEU-1/2 calculates percentage of unigram/up to bigram matching. DIST-1/2 calculates percentage of distinct unigram/bigram in generated responses. Reports of previous works use a different number of examples in training/test sets.}
    \begin{tabular}{l|ccc}
      \multicolumn{4}{c}{} \\
      \textbf{Method} & \textbf{BLEU-1/2} & \textbf{DIST-1/2} & \textbf{Length} \\ \hline
      Reference & 100/100 & 8.68/44.4 & 7.21 \\ \hline
      MMI \shortcite{li2016diversity} & - & 1.84/6.6 & - \\
      RL \shortcite{li2016deep} & - & 1.7/4.1 & - \\
      DP-GAN \shortcite{xu2018dp} & - & 2.39/11.1 & - \\ \hline
      Seq2Seq & 13.3/2.95 & 1.43/4.79 & 5.78 \\
      MemN2N & \textbf{13.6}/3.11 & 1.80/7.20 & 5.76 \\
      Seq2Seq + Attn & 13.3/\textbf{3.67}& 4.02/14.1 & 5.56 \\
      Seq2Seq + MMI & 12.2/2.53 & 6.54/\textbf{25.9} & 5.32 \\ \hline
      Seq2Seq + ITF & 12.9/2.70 & \textbf{7.56}/21.6 & \textbf{6.07} \\
    \end{tabular}
    \label{tab:opensubtitles}
  \end{center}
\end{table}

\begin{table}[t]
  \begin{center}
    \caption{Results of English and Japanese Twitter replies}
    \begin{tabular}{l|ccc}
      \multicolumn{4}{c}{} \\
      \multicolumn{4}{c}{\textbf{English}} \\
      \textbf{Method} & \textbf{BLEU-1/2} & \textbf{DIST-1/2} & \textbf{Length} \\ \hline
      Reference & 100/100 & 10.2/57.3 & 13.7 \\ \hline
      Seq2Seq & \textbf{10.6}/\textbf{3.25} & 1.25/5.66 & \textbf{11.3} \\
      Seq2Seq + MMI & 7.12/1.66 & 6.06/\textbf{33.0} & 9.24 \\ \hline
      Seq2Seq + ITF & 7.50/2.14 & \textbf{7.67}/26.3 & 10.3 \\
      \multicolumn{4}{c}{} \\
      \multicolumn{4}{c}{\textbf{Japanese}} \\
      \textbf{Method} & \textbf{BLEU-1/2} & \textbf{DIST-1/2} & \textbf{Length} \\ \hline
      Reference & 100/100 & 16.2/71.0 & 10.9 \\ \hline
      Seq2Seq & 10.7/2.63 & 7.98/26.3 & 6.11 \\
      MemN2N & \textbf{13.8}/\textbf{3.84} & 7.83/29.8 & 7.12 \\
      MemN2N + MMI & 12.6/2.69 & 14.5/\textbf{55.1} & 7.27 \\ \hline
      MemN2N + ITF & 12.8/3.03 & \textbf{16.8}/54.3 & \textbf{8.27} \\
    \end{tabular}
    \label{tab:twitter}
  \end{center}
\end{table}

We extracted dialogue data from the OpenSubtitles2018 corpus \cite{lison2016opensubtitles}.
This corpus has multiple subtitles for the same movie, but we used only one subtitle per movie to avoid an imbalanced training set.
In this corpus, we obtained the start and end times of each turn of the subtitles.
Each episode was configured as continuous turns with the interval from the end time of a turn to the start time of the next turn within five seconds.
As a result, the OpenSubtitles training set consists of 5M turns and 0.4M episodes (i.e., 4.6M examples).
Since all the episodes have multi-turns, we can use the memory network to consider the dialogue history.
The validation and test sets have 10k examples respectively.

%%%

Table \ref{tab:opensubtitles} shows that our Seq2Seq trained with ITF loss establishes a state-of-the-art DIST-1 of 7.56 while maintaining a good BLEU score.
Regarding the relative improvement of DIST-1 from the baseline Seq2Seq, MMI-antiLM \cite{li2016diversity} reported 228\%, RL \cite{li2016deep} reported 174\%, and DP-GAN \cite{xu2018dp} reported 25\%, but our ITF loss model achieved 429\%.
Seq2Seq with MMI inference increased DIST, but slightly decreased BLEU.
Seq2Seq with Attention increased BLEU-2 and DIST.
MemN2N achieved the highest BLEU-1 of 13.6, but its DIST improvement was small.

\subsubsection{Twitter}

We collected datasets of both English and Japanese Twitter replies.
We excluded self-replied dialogues, bot-to-bot dialogues, and extremely long dialogues from these data.
The English Twitter training set consists of 5M turns and 2.5M episodes (i.e., 2.5M examples).
All episodes have two turns.
The Japanese Twitter training set consists of 4.5M turns and 0.7M episodes (i.e., 3.8M examples).
All of the episodes have multi-turns.
On both the English and Japanese datasets, the validation and test sets have 10k examples respectively.

%%%

Table \ref{tab:twitter} shows that on both the English and Japanese datasets, our ITF loss model outperforms the MMI inference model on both BLEU-1 and DIST-1.
In particular, on the Japanese dataset, our loss model achieved a DIST-1 score of 16.8 compared to a ground truth of 16.2.

\subsection{Selection of $\lambda$ in ITF Loss}\label{sec:lambda}

We investigated the optimal value of hyperparameter $\lambda$ in Eq. \ref{eq:weight} through which the ITF loss model yields high diversity while maintaining good quality.
We used a set of $\lambda = \{0, 0.05, \ldots, 0.6, 0.65\}$ and trained Seq2Seq on an OpenSubtitles dialogue dataset that consists of 500k turns.

\begin{figure}[t]
  \begin{center}
    \includegraphics[width=80mm]{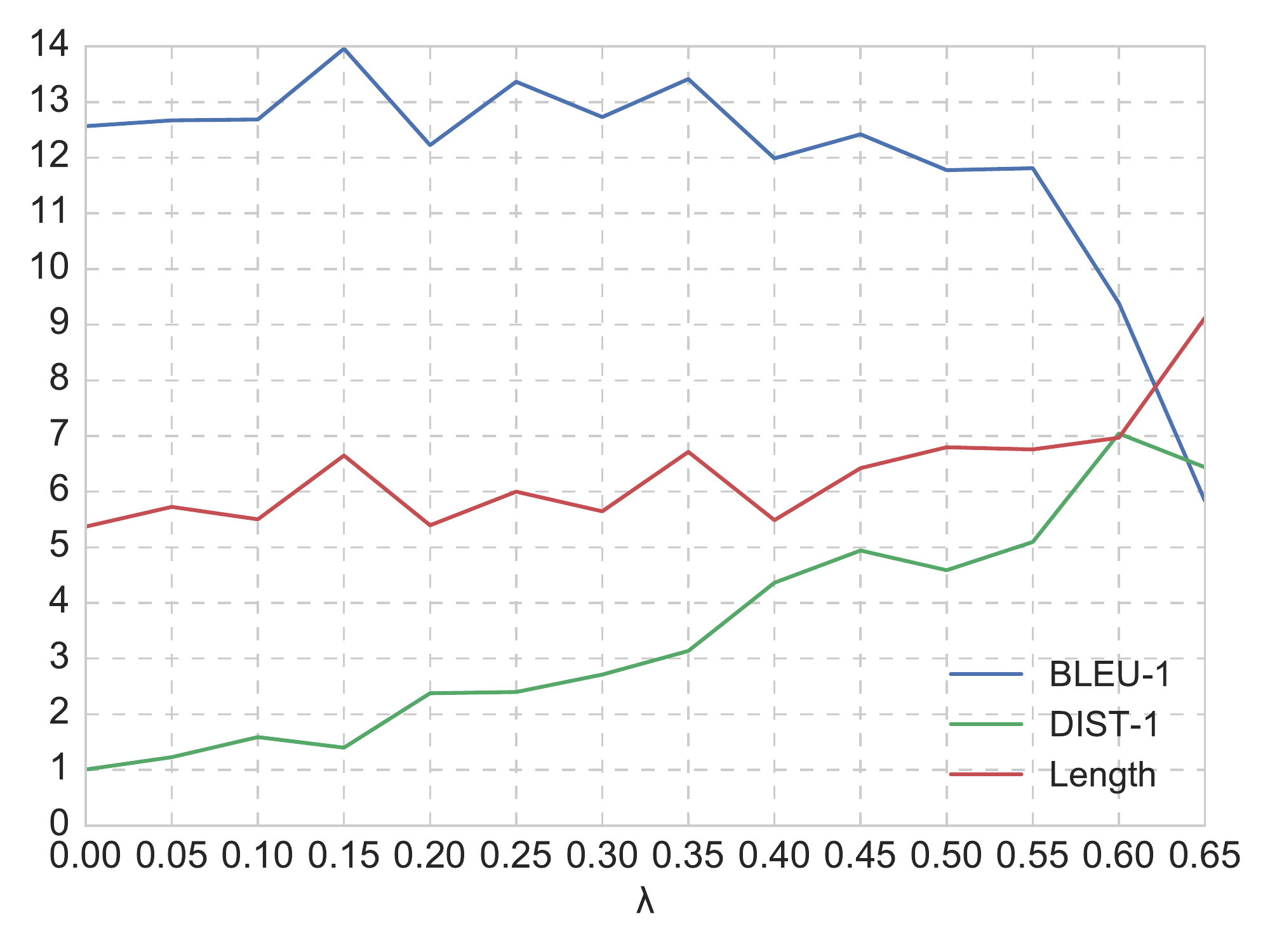}
    \caption{Comparison of automatic evaluation scores for each $\lambda$ in ITF loss on OpenSubtitles dialogue. Number of turns in training set is 500k, which is $1/10$ compared with Section \ref{sec:opensubtitles_data} and the number of subwords is 16k.}
    \label{fig:lambda}
  \end{center}
\end{figure}

Figure \ref{fig:lambda} shows the results. The generated sentences have a sufficiently high DIST-1 while maintaining a high BLEU-1 using $\lambda$ around $0.4$.

\subsection{ITF Inference in MLE Model}

\begin{table}[t]
  \begin{center}
    \caption{Comparison of automatic evaluation scores for each inference method. ITF infer is different from ITF loss.}
    \begin{tabular}{l|ccc}
      \multicolumn{4}{c}{} \\
      \textbf{Inference} & $\lambda$ & \textbf{BLEU-1} & \textbf{DIST-1} \\ \hline
      Noisy infer. & 1.4 & 12.5 & 2.81 \\
      MMI infer. & 0.7 & 12.5 & 4.76 \\
      ITF infer. & 0.09 & 12.5 & 4.85 \\
    \end{tabular}
    \label{tab:ITF_infer}
  \end{center}
\end{table}

We introduce another inference version of ITF loss, which applies the concept of inverse token frequency to the model trained with MLE during the evaluation. It resembles the use of MMI inference \cite{li2016diversity}.
One advantage is that it is unnecessary to re-run the training when we use different $\lambda$ values, compared to using ITF loss during the training.
ITF inference generates token $y$:

\begin{equation}
  y = {\rm argmax} \left\{\log {\rm softmax} (w \odot x) \right\},
\end{equation}

\noindent where $x \in \mathbb{R}^{|V|}$ is the output of the projection layer, $w \in \mathbb{R}^{|V|}$ is the weight (i.e., the vector version of Eq. \ref{eq:weight}), and $\odot$ is the element-wise product.

%%%

We also introduce a noisy inference to prove that the ITF and MMI inferences have more meaning than just noise injection:

\begin{equation}
  y = {\rm argmax} \left\{\log {\rm softmax} (x + \lambda {\rm noise}) \right\},
\end{equation}

\noindent where ${\rm noise} \in \mathbb{R}^{|V|}$ is the sampling from standard normal distribution $\mathcal{N}(0, 1)$.

%%%

Table \ref{tab:ITF_infer} shows that the performance of our ITF inference is close to MMI inference, and both are superior to noisy inference. We chose each $\lambda$ to be equivalent BLEU scores.

\subsection{Suppression of Repetitive Generation}\label{sec:sup_rep}

In our preliminary experiment, the decoder generated repetitive phrases (Table \ref{tab:sup_rep_examples}), which gravely decreased the quality of the generated responses.
This problem can be avoided by suppressing the regeneration of the already generated tokens during the decoding process.
We defined a repetition suppressor:

\begin{equation}
  {\rm suppressor}(x_k) = \frac{1}{\left\{1 + {\rm count}({\rm token}_k)\right\}^{\lambda}} x_k,
\label{eq:suppressor}
\end{equation}

\noindent where $x_k$ is the $k$-th element of $x \in \mathbb{R}^{|V|}$, which is the output of the projection layer and ${\rm count}({\rm token}_k)$ is the number of times ${\rm token}_k$ was generated in previous time steps during the decoding process.

\begin{table}[t]
  \begin{center}
    \caption{Examples of repetitive generation and its suppression}
    \begin{tabular}{l}
      \\ 
      \textbf{do nothing} \\
      i'm sorry to hear that. i'm sorry to hear that. \\
      i'm not sure i'm a cop. i'm not a cop. i'm not a cop. \\
      \textbf{suppress repetition} ($\lambda = 1$) \\
      i'm sorry to hear that. hope you enjoyed it! \\
      i'm not sure how you can do that. \\
    \end{tabular}
    \label{tab:sup_rep_examples}
  \end{center}
\end{table}

\begin{table}[t]
  \begin{center}
    \caption{Percentage of generated sentences containing identical tokens in all generated sentences. $\lambda = 0$ means that no repetition suppressor was used.}
    \begin{tabular}{l|ccc}
      \multicolumn{4}{c}{} \\
      \textbf{Dataset} & \multicolumn{3}{c}{\textbf{Repetition}} \\ \hline
      $\lambda$ & 0 & 0.5 & 1 \\ \hline
      OpenSubtitles & 6.22 & 1.08 & \textbf{0.93} \\
      English Twitter & 64.4 & 25.3 & \textbf{21.8} \\
      Japanese Twitter & 45.1 & 16.4 & \textbf{6.66} \\
    \end{tabular}
    \label{tab:sup_rep}
  \end{center}
\end{table}

We calculated the percentage of the generated sentences containing the same token in all the generated sentences.
Table \ref{tab:sup_rep} shows that the repetition suppressor avoids the regeneration of the previously generated tokens in all the datasets.

\section{Discussion}

Methods and results answer the following two questions about ITF loss.

\begin{quote}
1) Does ITF loss potentially generate grammatically incorrect sentences more often?
\end{quote}

Our generated responses in the OpenSubtitles dialogue (Table \ref{tab:opensubtitles_samples}) and the Twitter replies (Table \ref{tab:twitter_samples}, \ref{tab:japanese_twitter_samples}) show that the ITF loss model yields high diversity and quality that exceed or equal the result of the Maximum Mutual Information (MMI) inference model.
We found no serious grammatical problems in the generated sentences when it is less than $\lambda = 0.7$.

\begin{quote}
2) Does ITF loss modify the underlying distribution of tokens when generating a response?
\end{quote}

The goals of SCE and ITF loss are the same: getting the correct distribution of tokens.
ITF loss is not directly used to modify the distribution of tokens. It only applies to expected errors (a.k.a., SCE loss) from the distribution of tokens to avoid the overvaluation of frequent tokens.
In other words, scale $w$ in Eq. \ref{eq:weight}) just controls its contributions of the errors and ITF loss controls the gradients on individual tokens by frequency-based scaling.
This speeds up the learning on rare tokens and slows it down on frequent tokens.

\begin{table}[p]
  \begin{center}
    \caption{Generated responses on OpenSubtitles dialogue. SRC and TGT denote source and target sentences.}
    \begin{tabular}{p{8mm}p{65mm}}
      \\
      \textbf{SRC} & \textbf{Does he know what's going on?} \\
      \textbf{TGT} & \textbf{He knows he's leaving.} \\
      MLE & No. \\
      MMI & No. \\
      ITF & He's got a lot of trouble. \\
      \\ \hline
      \\
      \textbf{SRC} & \textbf{I'm not interested in talking to you.} \\
      \textbf{SRC} & \textbf{I'm leaving.} \\
      \textbf{TGT} & \textbf{Have the trial by yourselves.} \\
      MLE & I'm going to the bathroom. \\
      MMI & Okay. \\
      ITF & I'll be back tomorrow morning. \\
      \\ \hline
      \\
      \textbf{SRC} & \textbf{And you don't know Batman.} \\
      \textbf{SRC} & \textbf{He's not letting this lunatic just run wild.} \\
      \textbf{TGT} & \textbf{He can't catch him either.} \\
      MLE & You're not going to be a mutant. \\
      MMI & Then why he's not here? \\
      ITF & He's a hero. \\
      \\ \hline
      \\
      \textbf{SRC} & \textbf{My client, Turatello, is coming.} \\
      \textbf{SRC} & \textbf{There'll be chaos.} \\
      \textbf{SRC} & \textbf{They did a great job on your ass.} \\
      \textbf{TGT} & \textbf{Get well soon.} \\
      MLE & I'm sure you have a good friend. \\
      MMI & But you have no idea how much I'm going to get away. \\
      ITF & I was hoping to be honest, but you're not interested in the first time. \\
      \\ \hline
      \\
      \textbf{SRC} & \textbf{It's not loaded, look.} \\
      \textbf{SRC} & \textbf{See?} \\
      \textbf{SRC} & \textbf{We just need some help, that's all.} \\
      \textbf{TGT} & \textbf{Our friends, they're trapped.} \\
      MLE & I'm sorry. \\
      MMI & Fine, syringes. \\
      ITF & What do you want? \\
      \\ \hline
      \\
      \textbf{SRC} & \textbf{Sorry to leave you in the line.} \\
      \textbf{SRC} & \textbf{At the end of it, ok.} \\
      \textbf{SRC} & \textbf{They all cheat in Italy.} \\
      \textbf{TGT} & \textbf{Off you go.} \\
      MLE & What? \\
      MMI & They sell boats? \\
      ITF & They're not idiots who is in Italy, but we can make a lot of trouble.
    \end{tabular}
    \label{tab:opensubtitles_samples}
  \end{center}
\end{table}

\begin{table}[p]
  \begin{center}
    \caption{Generated responses on English Twitter replies. SRC and TGT denote source and target sentences.}
    \begin{tabular}{p{8mm}p{65mm}}
      \\
      \textbf{SRC} & \textbf{i hope you had a wonderful trip! can that flight please pick me up to go to iceland? what a beautiful place!} \\
      \textbf{TGT} & \textbf{place it on your short list of ``to-do's.''. it's a beautiful country.} \\
      MLE & i'll be back in december! \\
      MMI & poutine flight iceland has a great time. \\
      ITF & yes! iceland is awesome :) \\
      \\ \hline
      \\
      \textbf{SRC} & \textbf{i don't think it'll happen, but apple does have the money..} \\
      \textbf{TGT} & \textbf{i suspect the design teams are fundamentally incompatible. mclaren much more radical and more performance oriented.} \\
      MLE & i'm not sure if it is, but the fact that the android phone is a bad thing. \\
      MMI & true apple products are the best. \\
      ITF & i think apple is working on android aosp, pixel phones are very good. \\
      \\ \hline
      \\
      \textbf{SRC} & \textbf{12 gb ram at 384 gb/sec (gddr5x).. if this is true than damnn!.. you want} \\
      \textbf{TGT} & \textbf{eurogamer also said that it speculates 384 and 12 gb of ram by placement of ram in original video} \\
      MLE & i'm not sure if it is worth it. \\
      MMI & mwr gpu is the best. \\
      ITF & rambo is a newer one with chromebook7 connector, laptop router, hdmi cables. \\
      \\ \hline
      \\
      \textbf{SRC} & \textbf{10 reasons you need to exercise with your bodyweight} \\
      \textbf{TGT} & \textbf{yep, planks, side planks, etc are my go to!} \\
      MLE & i'm not a good person, but it is hard to get through. \\
      MMI & metabolism thighs lotion depending on how you feel \\
      ITF & 10 yrs later: 1.15 pounds of weight rate increases so much easier than a lot. \\
      \\ \hline
      \\
      \textbf{SRC} & \textbf{im going back to dragon ball z, while yall talking about pikachu, ill be talking about goku} \\
      \textbf{TGT} & \textbf{which goku} \\
      MLE & i'm not even gonna be able to play that game \\
      MMI & vegeta and goku literally \\
      ITF & i knoww, goku rap eminems remix \\
    \end{tabular}
    \label{tab:twitter_samples}
  \end{center}
\end{table}

\begin{table}[p]
  \begin{center}
    \caption{Generated responses on Japanese Twitter replies. SRC and TGT denote source and target sentences. We manually replaced emoticon and emoji to similar Font Awesome.}
    \begin{tabular}{c}
      \includegraphics[width=81mm]{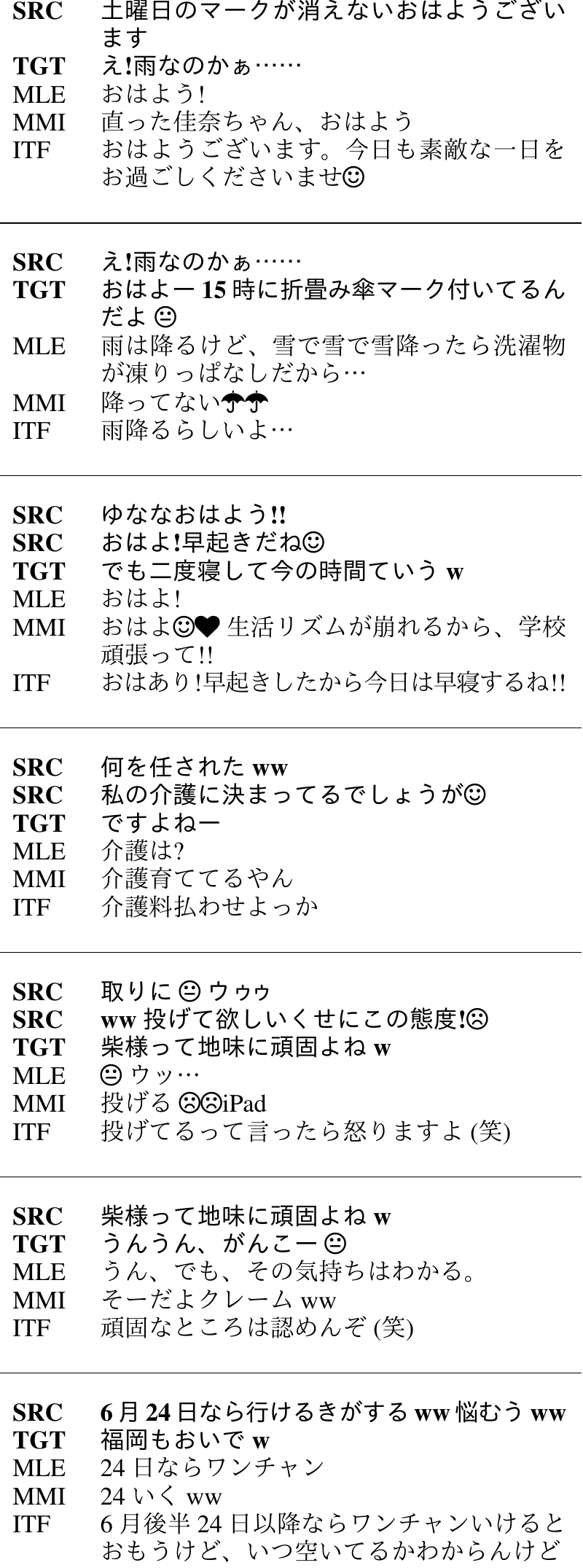}
    \end{tabular}
    \label{tab:japanese_twitter_samples}
  \end{center}
\end{table}

\begin{table}[p]
  \begin{center}
    \begin{tabular}{c}
      \includegraphics[width=81mm]{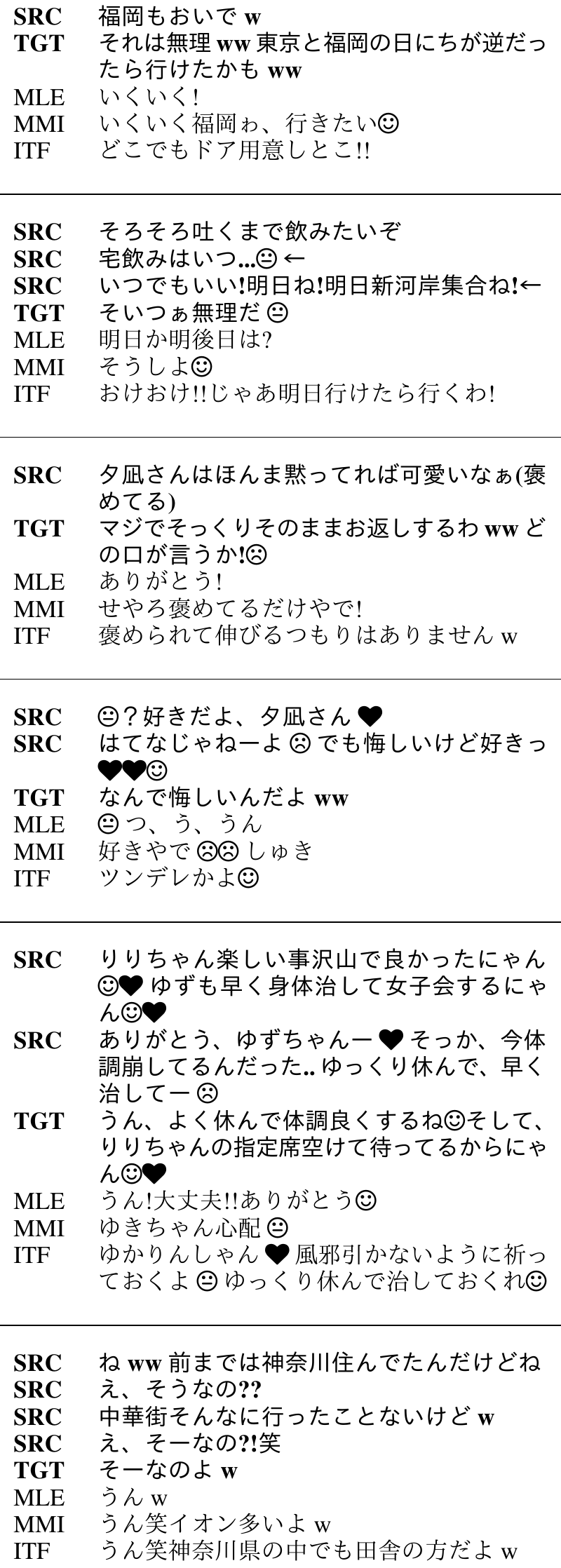}
    \end{tabular}
    \label{tab:japanese_twitter_samples_2}
  \end{center}
\end{table}

\section{Conclusion}

We focused on the low diversity problem and confirmed that unigram diversity scores significantly improve by applying Inverse Token Frequency loss.
Future work will investigate Inverse N-gram Frequency (INF) loss, which is a general type of ITF loss that only considers unigram frequency.
Since BLEU is not well suited to evaluate dialogue responses, we are planning to conduct human evaluations.

\bibliographystyle{aaai}
\bibliography{bibfile}

\begin{thebibliography}{}

\bibitem[\protect\citeauthoryear{Bowman \bgroup et al\mbox.\egroup
  }{2016}]{bowman2016generating}
Bowman, S.~R.; Vilnis, L.; Vinyals, O.; Dai, A.; Jozefowicz, R.; and Bengio, S.
\newblock 2016.
\newblock Generating sentences from a continuous space.
\newblock In {\em Proceedings of The 20th SIGNLL Conference on Computational
  Natural Language Learning},  10--21.

\bibitem[\protect\citeauthoryear{Cao and Clark}{2017}]{cao2017latent}
Cao, K., and Clark, S.
\newblock 2017.
\newblock Latent variable dialogue models and their diversity.
\newblock In {\em Proceedings of the 15th Conference of the European Chapter of
  the Association for Computational Linguistics (Volume 2, Short Papers)},
  volume~2,  182--187.

\bibitem[\protect\citeauthoryear{Ghazvininejad \bgroup et al\mbox.\egroup
  }{2017}]{ghazvininejad2017knowledge}
Ghazvininejad, M.; Brockett, C.; Chang, M.-W.; Dolan, B.; Gao, J.; Yih, W.-t.;
  and Galley, M.
\newblock 2017.
\newblock A knowledge-grounded neural conversation model.
\newblock {\em arXiv preprint arXiv:1702.01932}.

\bibitem[\protect\citeauthoryear{Goodfellow \bgroup et al\mbox.\egroup
  }{2014}]{goodfellow2014generative}
Goodfellow, I.; Pouget-Abadie, J.; Mirza, M.; Xu, B.; Warde-Farley, D.; Ozair,
  S.; Courville, A.; and Bengio, Y.
\newblock 2014.
\newblock Generative adversarial nets.
\newblock In {\em Advances in neural information processing systems},
  2672--2680.

\bibitem[\protect\citeauthoryear{Kingma and Welling}{2013}]{kingma2013auto}
Kingma, D.~P., and Welling, M.
\newblock 2013.
\newblock Auto-encoding variational bayes.
\newblock {\em arXiv preprint arXiv:1312.6114}.

\bibitem[\protect\citeauthoryear{Kudo}{2018}]{kudo2018subword}
Kudo, T.
\newblock 2018.
\newblock Subword regularization: Improving neural network translation models
  with multiple subword candidates.
\newblock {\em arXiv preprint arXiv:1804.10959}.

\bibitem[\protect\citeauthoryear{Li \bgroup et al\mbox.\egroup
  }{2016a}]{li2016diversity}
Li, J.; Galley, M.; Brockett, C.; Gao, J.; and Dolan, B.
\newblock 2016a.
\newblock A diversity-promoting objective function for neural conversation
  models.
\newblock In {\em Proceedings of the 2016 Conference of the North American
  Chapter of the Association for Computational Linguistics: Human Language
  Technologies},  110--119.

\bibitem[\protect\citeauthoryear{Li \bgroup et al\mbox.\egroup
  }{2016b}]{li2016deep}
Li, J.; Monroe, W.; Ritter, A.; Galley, M.; Gao, J.; and Jurafsky, D.
\newblock 2016b.
\newblock Deep reinforcement learning for dialogue generation.
\newblock In {\em Proceedings of the 2016 Conference on Empirical Methods in
  Natural Language Processing},  1192--1202.

\bibitem[\protect\citeauthoryear{Li \bgroup et al\mbox.\egroup
  }{2017}]{li2017adversarial}
Li, J.; Monroe, W.; Shi, T.; Jean, S.; Ritter, A.; and Jurafsky, D.
\newblock 2017.
\newblock Adversarial learning for neural dialogue generation.
\newblock In {\em Proceedings of the 2017 Conference on Empirical Methods in
  Natural Language Processing},  2157--2169.

\bibitem[\protect\citeauthoryear{Lison and
  Tiedemann}{2016}]{lison2016opensubtitles}
Lison, P., and Tiedemann, J.
\newblock 2016.
\newblock Opensubtitles2016: Extracting large parallel corpora from movie and
  tv subtitles.
\newblock In {\em Proceedings of the 10th International Conference on Language
  Resources and Evaluation}.

\bibitem[\protect\citeauthoryear{Miller \bgroup et al\mbox.\egroup
  }{2016}]{miller2016key}
Miller, A.; Fisch, A.; Dodge, J.; Karimi, A.-H.; Bordes, A.; and Weston, J.
\newblock 2016.
\newblock Key-value memory networks for directly reading documents.
\newblock In {\em Proceedings of the 2016 Conference on Empirical Methods in
  Natural Language Processing},  1400--1409.

\bibitem[\protect\citeauthoryear{Olabiyi \bgroup et al\mbox.\egroup
  }{2018}]{olabiyi2018multi}
Olabiyi, O.; Salimov, A.; Khazane, A.; and Mueller, E.
\newblock 2018.
\newblock Multi-turn dialogue response generation in an adversarial learning
  framework.
\newblock {\em arXiv preprint arXiv:1805.11752}.

\bibitem[\protect\citeauthoryear{Papineni \bgroup et al\mbox.\egroup
  }{2002}]{papineni2002bleu}
Papineni, K.; Roukos, S.; Ward, T.; and Zhu, W.-J.
\newblock 2002.
\newblock Bleu: a method for automatic evaluation of machine translation.
\newblock In {\em Proceedings of the 40th Annual Meeting of the Association for
  Computational Linguistics},  311--318.

\bibitem[\protect\citeauthoryear{Serban \bgroup et al\mbox.\egroup
  }{2016}]{serban2016building}
Serban, I.~V.; Sordoni, A.; Bengio, Y.; Courville, A.; and Pineau, J.
\newblock 2016.
\newblock Building end-to-end dialogue systems using generative hierarchical
  neural network models.
\newblock In {\em Thirtieth AAAI Conference on Artificial Intelligence}.

\bibitem[\protect\citeauthoryear{Serban \bgroup et al\mbox.\egroup
  }{2017a}]{serban2017deep}
Serban, I.~V.; Sankar, C.; Germain, M.; Zhang, S.; Lin, Z.; Subramanian, S.;
  Kim, T.; Pieper, M.; Chandar, S.; Ke, N.~R.; et~al.
\newblock 2017a.
\newblock A deep reinforcement learning chatbot.
\newblock {\em arXiv preprint arXiv:1709.02349}.

\bibitem[\protect\citeauthoryear{Serban \bgroup et al\mbox.\egroup
  }{2017b}]{serban2017hierarchical}
Serban, I.~V.; Sordoni, A.; Lowe, R.; Charlin, L.; Pineau, J.; Courville, A.;
  and Bengio, Y.
\newblock 2017b.
\newblock A hierarchical latent variable encoder-decoder model for generating
  dialogues.
\newblock In {\em Thirty-First AAAI Conference on Artificial Intelligence}.

\bibitem[\protect\citeauthoryear{Shang, Lu, and Li}{2015}]{shang2015neural}
Shang, L.; Lu, Z.; and Li, H.
\newblock 2015.
\newblock Neural responding machine for short-text conversation.
\newblock In {\em Proceedings of the 53rd Annual Meeting of the Association for
  Computational Linguistics and the 7th International Joint Conference on
  Natural Language Processing (Volume 1: Long Papers)}, volume~1,  1577--1586.

\bibitem[\protect\citeauthoryear{Shao \bgroup et al\mbox.\egroup
  }{2017}]{shao2017generating}
Shao, Y.; Gouws, S.; Britz, D.; Goldie, A.; Strope, B.; and Kurzweil, R.
\newblock 2017.
\newblock Generating high-quality and informative conversation responses with
  sequence-to-sequence models.
\newblock In {\em Proceedings of the 2017 Conference on Empirical Methods in
  Natural Language Processing},  2210--2219.

\bibitem[\protect\citeauthoryear{Sordoni \bgroup et al\mbox.\egroup
  }{2015}]{sordoni2015neural}
Sordoni, A.; Galley, M.; Auli, M.; Brockett, C.; Ji, Y.; Mitchell, M.; Nie,
  J.-Y.; Gao, J.; and Dolan, B.
\newblock 2015.
\newblock A neural network approach to context-sensitive generation of
  conversational responses.
\newblock In {\em Proceedings of the 2015 Conference of the North American
  Chapter of the Association for Computational Linguistics: Human Language
  Technologies},  196--205.

\bibitem[\protect\citeauthoryear{Sukhbaatar \bgroup et al\mbox.\egroup
  }{2015}]{sukhbaatar2015end}
Sukhbaatar, S.; Weston, J.; Fergus, R.; et~al.
\newblock 2015.
\newblock End-to-end memory networks.
\newblock In {\em Advances in neural information processing systems},
  2440--2448.

\bibitem[\protect\citeauthoryear{Vaswani \bgroup et al\mbox.\egroup
  }{2017}]{vaswani2017attention}
Vaswani, A.; Shazeer, N.; Parmar, N.; Uszkoreit, J.; Jones, L.; Gomez, A.~N.;
  Kaiser, {\L}.; and Polosukhin, I.
\newblock 2017.
\newblock Attention is all you need.
\newblock In {\em Advances in Neural Information Processing Systems},
  5998--6008.

\bibitem[\protect\citeauthoryear{Vinyals and Le}{2015}]{vinyals2015neural}
Vinyals, O., and Le, Q.
\newblock 2015.
\newblock A neural conversational model.
\newblock {\em arXiv preprint arXiv:1506.05869}.

\bibitem[\protect\citeauthoryear{Wen \bgroup et al\mbox.\egroup
  }{2015a}]{wen2015stochastic}
Wen, T.-H.; Ga{\v{s}}ic, M.; Kim, D.; Mrk{\v{s}}ic, N.; Su, P.-H.; Vandyke, D.;
  and Young, S.
\newblock 2015a.
\newblock Stochastic language generation in dialogue using recurrent neural
  networks with convolutional sentence reranking.
\newblock In {\em 16th Annual Meeting of the Special Interest Group on
  Discourse and Dialogue},  275.

\bibitem[\protect\citeauthoryear{Wen \bgroup et al\mbox.\egroup
  }{2015b}]{wen2015semantically}
Wen, T.-H.; Gasic, M.; Mrksic, N.; Su, P.-H.; Vandyke, D.; and Young, S.
\newblock 2015b.
\newblock Semantically conditioned lstm-based natural language generation for
  spoken dialogue systems.
\newblock In {\em Proceedings of the 2015 Conference on Empirical Methods in
  Natural Language Processing},  1711--1721.

\bibitem[\protect\citeauthoryear{Xu \bgroup et al\mbox.\egroup
  }{2018}]{xu2018dp}
Xu, J.; Sun, X.; Ren, X.; Lin, J.; Wei, B.; and Li, W.
\newblock 2018.
\newblock Dp-gan: Diversity-promoting generative adversarial network for
  generating informative and diversified text.
\newblock {\em arXiv preprint arXiv:1802.01345}.

\bibitem[\protect\citeauthoryear{Yu \bgroup et al\mbox.\egroup
  }{2017}]{yu2017seqgan}
Yu, L.; Zhang, W.; Wang, J.; and Yu, Y.
\newblock 2017.
\newblock Seqgan: Sequence generative adversarial nets with policy gradient.
\newblock In {\em Thirty-First AAAI Conference on Artificial Intelligence}.

\bibitem[\protect\citeauthoryear{Zhao, Zhao, and
  Eskenazi}{2017}]{zhao2017learning}
Zhao, T.; Zhao, R.; and Eskenazi, M.
\newblock 2017.
\newblock Learning discourse-level diversity for neural dialog models using
  conditional variational autoencoders.
\newblock In {\em Proceedings of the 55th Annual Meeting of the Association for
  Computational Linguistics (Volume 1: Long Papers)}, volume~1,  654--664.

\bibitem[\protect\citeauthoryear{Zhou \bgroup et al\mbox.\egroup
  }{2017}]{zhou2017mechanism}
Zhou, G.; Luo, P.; Cao, R.; Lin, F.; Chen, B.; and He, Q.
\newblock 2017.
\newblock Mechanism-aware neural machine for dialogue response generation.
\newblock In {\em Thirty-First AAAI Conference on Artificial Intelligence}.

\end{thebibliography}

\end{document}